\pdfoutput=1

\documentclass[11pt]{article}

\usepackage[]{ACL2023}

\usepackage{times}
\usepackage{latexsym}

\usepackage[T1]{fontenc}

\usepackage[utf8]{inputenc}

\usepackage{microtype}

\usepackage{inconsolata}
\usepackage{hyperref}
\usepackage{amsfonts}
\usepackage{booktabs}
\usepackage{multirow}
\usepackage{amstext}
\usepackage{bm}
\usepackage{cite} 
\usepackage{amsmath}
\usepackage{graphicx}
\usepackage{subfigure}
\usepackage{xcolor}
\usepackage{CJKutf8}
\usepackage{diagbox}
\usepackage{cleveref}

\title{A Neural Divide-and-Conquer Reasoning Framework \\ for Image Retrieval from Linguistically Complex Text}

\author{Yunxin Li$^{1}$\thanks{~~$^{\dagger}$Corresponding author.}, Baotian Hu$^{1\dagger}$, Yuxin Ding$^{1}$, Lin Ma$^{2}$, Min Zhang$^{1}$\\
$^{1}$Harbin Institute of Technology, Shenzhen, China, $^{2}$Meituan, Beijing\\
\texttt{\{hubaotian, yxding, zhangmin2021\}}@hit.edu.cn
\\
\texttt{liyunxin987@163.com}, 
\texttt{forest.linma@gmail.com}
}

\begin{document}

\maketitle

\begin{abstract}

    Pretrained Vision-Language Models (VLMs) have achieved remarkable performance in image retrieval from text. However, their performance drops drastically when confronted with linguistically complex texts that they struggle to comprehend.
    Inspired by the Divide-and-Conquer~\citep{smith1985design} algorithm and dual-process theory~\citep{groves1970habituation}, in this paper, we regard linguistically complex texts as compound proposition texts composed of multiple simple proposition sentences and propose an end-to-end Neural Divide-and-Conquer Reasoning framework, dubbed NDCR. It contains three main components: 1) \textit{Divide}: a proposition generator divides the compound proposition text into simple proposition sentences and produces their corresponding representations, 2) \textit{Conquer}: a pretrained VLMs-based visual-linguistic interactor achieves the interaction between decomposed proposition sentences and images, 3) \textit{Combine}: a neural-symbolic reasoner combines the above reasoning states to obtain the final solution via a neural logic reasoning approach. According to the dual-process theory, the visual-linguistic interactor and neural-symbolic reasoner could be regarded as analogical reasoning System 1 and logical reasoning System 2. We conduct extensive experiments on a challenging image retrieval from contextual descriptions data set. Experimental results and analyses indicate NDCR significantly improves performance in the complex image-text reasoning problem. Code link: \url{https://github.com/YunxinLi/NDCR}.

\end{abstract}

\section{Introduction}

\begin{figure}[t]
    \centering
    \includegraphics[width=0.45\textwidth]{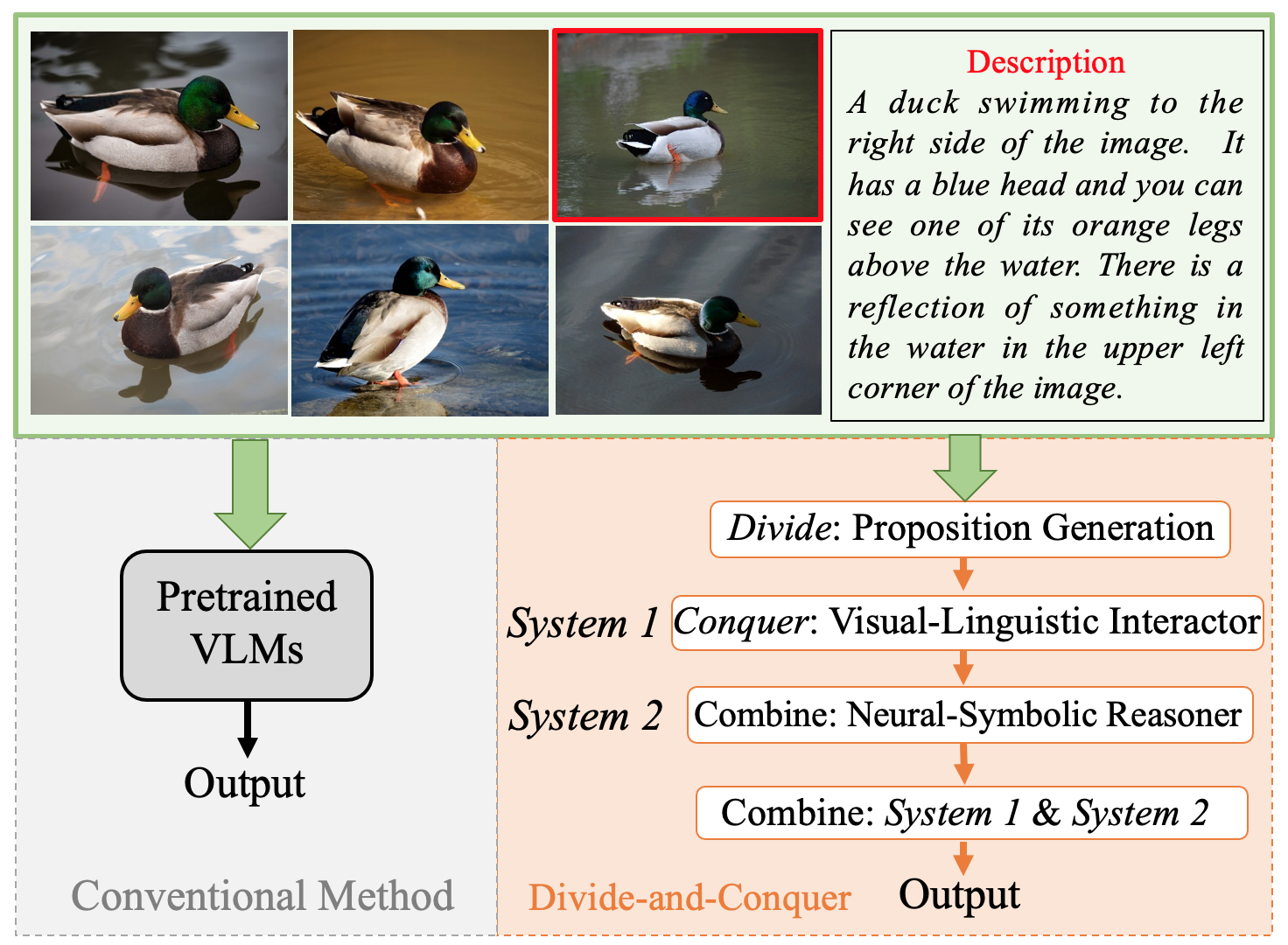}
    \caption{An example from the I{\footnotesize MAGE}C{\footnotesize O}D{\footnotesize E}~\citep{imagescode} data set, where the description is linguistically complex and images are minimally contrastive. The target image is in red and others are incorrect frames. The bottom part depicts the conventional method and the neural divide-and-conquer reasoning framework.}
    \label{fig:figure_1}
\end{figure}

Image-text retrieval tasks have made remarkable progress owing to pretrained Vision-Language Models (VLMs) such as LXMERT~\citep{tan-bansal-2019-lxmert}, UNITER~\citep{chen2020uniter}, OSCAR~\citep{li2020oscar, zhang2021vinvl}, ViLBERT~\citep{vilbert}, CLIP~\citep{CLIP}, and many others. These VLMs are usually trained on the large-scale short text-image corpus by cross-modal semantic alignment methods. They are capable of essential perceptual computing capability and excel at retrieving images from sentences with few objects and simple linguistic, e.g., \textit{``There is a duck swimming in the pond''}. However, when pretrained VLMs meet the case of retrieving the accurate image from similar candidates based on a linguistically complex text, as the example shown in Figure~\ref{fig:figure_1}, previous works~\citep{imagescode, Talmor2021CommonsenseQA2E, thrush2022winoground} show that they struggle to understand the elaborate description and perform complex cross-modal reasoning.

According to the dual-process theory for human thinking~\citep{groves1970habituation, evans2003two, pelaccia2011analysis}, human brains contain two thinking systems: System 1 performs analogical reasoning well, which is fast yet unconscious; System 2 is capable of abstract logical reasoning, which is slow yet conscious and well-suitable for complex reasoning problems. The theory could also hold for the image-text retrieval tasks, and the widely adopted models (e.g., VLMs) focus on analogical reasoning as System 1 based on the analysis of deep learning networks~\citep{bengio2017consciousness, system_1_2, bengio2021deep}. For the linguistically complex description that contains multiple conditions, they have inferior performance, and we need to introduce logical reasoning System 2 more to cover and logically incorporate the scattered information in the description based on System 1. Inspired by the above investigations and classical Divide-and-Conquer~\citep{smith1985design} algorithm, we design an end-to-end Neural Divide-and-Conquer Reasoning framework named NDCR. As shown in Figure~\ref{fig:figure_1}, our key idea is to regard the complex description as compound proposition text and solve the challenging retrieval problem in three steps: divide, conquer, and combine.

Specifically, \textbf{Divide:} NDCR first utilizes a proposition generator to divide the complex compound text and produce the global representation of simple proposition sentences with visually printing them. \textbf{Conquer:} we devise a visual-linguistic interactor to achieve the interaction between decomposed proposition sentences and images, which resembles System 1. It uses the Transformer~\citep{attention}-based contextual interactor to achieve the inter-learning of different proposition-image pairs. Considering the incorrectness or information loss of simple proposition representation, we also present a modifier to incorporate the context reasoning information to improve their cross-modal reasoning states. \textbf{Combine:} we design a learnable neural-symbolic reasoner to integrate reasoning information of simple propositions logically. It first employs a negation executor to obtain a simple proposition sentence's negational reasoning hidden state and corresponding confidence score. Then, we use the global reasoning information of compound proposition text as the query signal to perform the conjunction operation across simple propositions and their negational information. Finally, as shown in Figure~\ref{fig:figure_1}, we also combine the inferred results of the neural-symbolic reasoner (resembles System 2) and visual-linguistic interactor (resembles System 1) to obtain the final solution. In this way, the whole framework integrate the capabilities of Systems 1 and 2 to obtain better performance.

We conduct extensive experiments on a large-scale image retrieval from contextual descriptions data set, I{\footnotesize MAGE}C{\footnotesize O}D{\footnotesize E}~\citep{imagescode}. The experimental results indicate that NDCR achieves the state-of-the-art performance and the ablation and case studies verify the effectiveness of different modules. 

Our contributions are as follows:
\begin{itemize}

    \item We propose a divide-and-conquer reasoning framework for image retrievals from linguistically complex text, where we first attempt to combine the perceptually analogical reasoning System 1 and neural-symbolic logic reasoning System 2 to solve the complex multi-modal reasoning problem.
    
    \item We design a proposition generator capable of producing the global representation of decomposed simple proposition sentences for linguistically complex texts and visually printing them as text.

    \item Experimental results indicate our approach remarkably improves the performance, and we obtain the first place on the leaderboard~\footnote{\url{https://mcgill-nlp.github.io/imagecode} }. Ablation and case studies confirm the effectiveness of introducing and combining logical reasoning System 2 based on System 1.
    
\end{itemize}

\begin{figure*}[t]
    \centering
    \includegraphics[width=0.93\textwidth, height=0.41\textwidth]{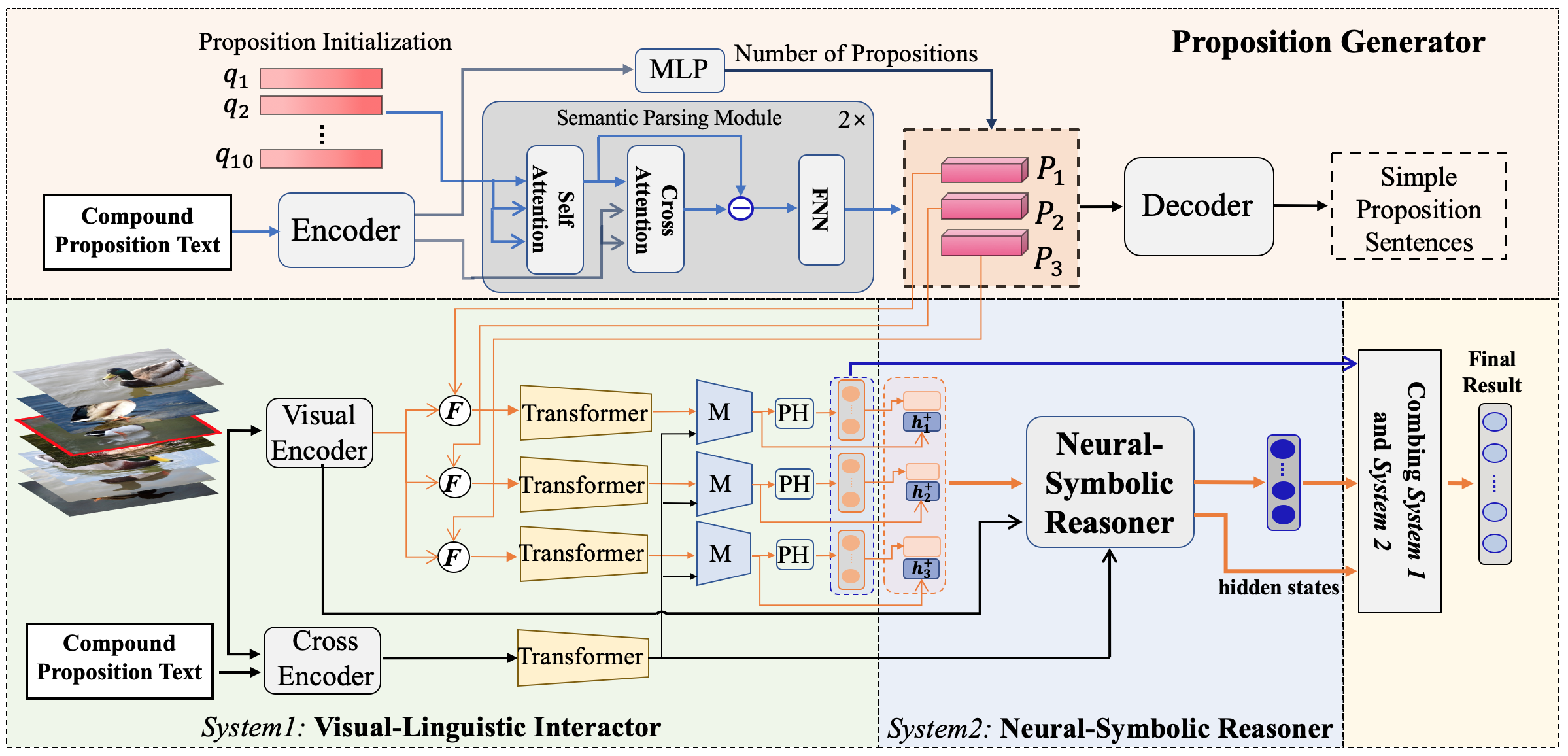}
    \caption{The overall architecture of neural divide-and-conquer reasoning framework. }
    \label{fig:model}
\end{figure*}

\section{Related Works}

\textbf{Pretrained Vision-Language Models for Cross Modal Matching.} 
Owing to the success of Transformer~\citep{attention} architecture equipped with pretrain-finetuning~\citep{pretrain-fintune} learning method, pretrained VLMs have made a remarkable performance in cross-modal matching or reasoning tasks~\citep{talmor2021multimodalqa}, especially image-text retrieval. Early pretrained VLMs utilize BERT~\citep{devlin-etal-2019-bert}-like single encoder architecture to encode and fuse the image-text information, then perform image-text reasoning such as ViLBERT~\citep{vilbert}, VisualBERT~\citep{li2019visualbert}, and Oscar~\citep{li2020oscar}. In addition, dual-encoder architecture such as CLIP~\citep{CLIP}, and ALBERT~\citep{ALBEF}, performs better than single-encoder architecture on image-text matching tasks and is widely used in industry because of its efficiency.

\noindent\textbf{Divide-and-Conquer for Question Answering.} The divide-and-conquer algorithm~\citep{smith1985design} aims to divide the complex problem into multiple simple problems and then combine the sub-problem results to achieve the final solution. This idea has been used in complex question-answering tasks in the natural language processing area. \citet{zhang-etal-2019-complex} proposed to utilize the decomposition of complex questions for semantic parsing. \citet{min-etal-2019-multi} adopt the question decomposition and rescoring method to perform multi-hop reading comprehension, which makes the reasoning path interpretable and robust. \citet{wolfson-etal-2022-weakly} utilized the QDMR structures of complex questions to conduct the decompose-synthesize text-to-SQL transformation. 
Previous pipeline approaches may lead to error cascades in the upper inference process due to the incompleteness or error of decomposed text. The image-text retrieval task has strict requirements on the correctness of text semantic understanding, thus we propose an end-to-end divide-and-conquer method for alleviating the error cascade issue via the whole learning process.

 \noindent\textbf{Dual-Process Theory.} 
The dual-process theory shows that human brains have two different thinking Systems. System 1 performs analogical reasoning, and System 2 performs conscious logical reasoning. Combining this theory with practical tasks, some researchers designed various approaches. \citet{mittal2017thinking} believed that combining vector space models with external knowledge graphs could be regarded as thinking `fast' in vector space along with thinking 'slow' and `deeply' by reasoning over the knowledge graph. \citet{anthony2017thinking} also proposed to use a deep learning network with a tree search engine as System 1 and System 2, respectively, for sequential decision-making problems. \citet{bengio2017consciousness, system_1_2} advocated the design of a conscious network to achieve the leap from System 1 to System 2. \citet{liu2022neural} designed a neural-symbolic system for natural language understanding tasks, which combines the explicit symbolic calculation-based System 2 and fast deep learning network-based System 1. 
For complex multi-modal reasoning problem, e.g., image retrieval from linguistically complex text, humans usually combine System 1 and System 2 to obtain the final solution. However, current methods relying mainly on deep learning networks resemble System 1 and lack the logical reasoning capability, thus suffering from image-text reasoning with the complex description. In this light, we make the first attempt to combine System 1 and System 2 to tackle this issue by designing a neural divide-and-conquer reasoning framework. We introduce a neural-symbolic reasoner in System 2 to conduct the logical operation. The overall framework contains analogical and logical reasoning as humans think, making appreciable gains.    

\section{Method}

\subsection{Overview}
Image retrieval from contextual descriptions~\citep{imagescode} aims to infer the correct image given a linguistically complex text $Y = (y_1, ..., y_N)$ and similar images $I = (I_1, ..., I_{L})$, where $y_i$, $N$, $I_i$, and $L$ represent the $i$ th token, the total length of text, $i$ th image, and the number of images, respectively.
We propose a novel divide-and-conquer reasoning framework to tackle such a task. It consists of three components, namely, Proposition Generator, Visual-Linguistic Interactor, and Neural-Symbolic Reasoner, which are coupled and trained in an end-to-end manner. Specifically, the proposition generator divides the complex description into multiple proposition sentences, allowing it to convert the complex matching problem to simple ones. Afterwards, the visual-linguistic interactor achieves the interaction between decomposed proposition sentences and images, resembling System 1, to perform the essential analogical reasoning. Subsequently, the neural-symbolic reasoner that relies on the reasoning state output by the visual-linguistic interactor resembles System 2 to perform logical reasoning. Finally, we also combine the output results of System 1 and System 2 to obtain the final solution.


\subsection{Proposition Generator}
The proposition generator is a sequence-to-sequence model based on the pretrained language model BART. As shown in Figure~\ref{fig:model}, it employs the encoder to obtain the text representation $\mathbf{H}_{Y} = (h_{cls}, h_{y_1}, ..., h_{y_N})$ where $h_{y_i}$ represents the $i$ th token hidden state. Subsequently, we design a two-layer semantic parsing module to gain the global representation of simple proposition sentences. Concretely, we set the maximum number of simple propositions to 10 and randomly initialize them. The initial vectors are fed to the semantic parsing module to interact with the compound text representation. Take the first layer as an example; the calculation process is following,
\begin{equation}
\begin{array}{c}
    \mathbf{h}^{T}_{s} = \text{Self-Attention}(\mathbf{h}^{I})\vspace{1.0ex},\\
    \mathbf{h}^{T}_{c} = \text{Cross-Attention}(\mathbf{h}^{T}_{s}, \mathbf{H}_Y)\vspace{1.0ex},\\
    \mathbf{h}^{T}_{F} = \text{FNN}(\mathbf{h}^{T}_{c} -  \mathbf{h}^{T}_{s}),
\end{array}
\label{eq1}
\end{equation}
where $\mathbf{h}^{I}$ is the randomly initial proposition representations. Attention and FNN calculation sub-networks are identical to the transformer~\citep{attention} architecture. Different from the transformer, we let the output of Cross-Attention layer subtract the output of Self-Attention layer, aiming to achieve information differences across propositions. 

By doing the same two-layer calculation, we obtain ten global hidden states of simple propositions. Due to context containing different numbers of simple proposition, we use a MLP to predict the target number of simple proposition sentences. It only attends to the global hidden state $h_{cls}$ of compound proposition text. Suppose that the predicted number $M$ of simple propositions is 3 (same as Figure~\ref{fig:model}), we adopt the first-three hidden states of the semantic parsing module as the global representation of the targeted simple proposition. As shown in Figure~\ref{fig:model}, for explaining what simple propositions represent, we also use the decoder of BART to generate the simple proposition sentence with only attending to their global representations. 


\subsection{System 1: Visual-Linguistic Interactor}
After obtaining the global representations of simple proposition sentences, we introduce the visual-linguistic interactor to mine the interaction of image-proposition pairs. Specifically, we use a pretrained visual encoder to obtain the image encoding representations $\mathbf{H}_{I} = (\mathbf{h}_{I_1}, ..., \mathbf{h}_{I_{L}})$ and fuse them with the simple proposition representation via the dot-product way (as the \textit{``F''} shown in Figure~\ref{fig:model}). The two-modal fusion process is 
$\mathbf{H}(p) = \lambda \cdot \text{Norm}(\mathbf{P}) \cdot \text{Norm}(\mathbf{H}_{I}),$ where $\lambda$ is the hyperparameter set to enlarge the scale of fused vectors. We denote the fused sequence representation of proposition-image pairs to $\mathbf{H}(p) = (\mathbf{H}(p_1), ..., \mathbf{H}(p_M))$ where $\mathbf{H}(p_1)$ indicates the sequential representation of first proposition combined with images. 

Then, we employ a two-layer transformer to perform the contextual information interaction for fused sequential representations $\mathbf{H}(p)$ and obtain the initial reasoning states of simple proposition on images. Considering the incorrectness or information loss of simple proposition representation obtained by the proposition generator, we introduce a MLP-based modifier to incorporate the reasoning state of compound proposition text to enhance previous initial reasoning states of simple propositions. The whole process is performed as Eq.~\ref{eq3},
\begin{equation}
\begin{array}{c}
    \mathbf{H}^{S_1}_{P} = \text{Transformer}(\mathbf{H}(p) + PE)\vspace{1.0ex},\\
    \mathbf{H}^{sg}_{C} = \text{Transformer}(\mathbf{H}_C + PE)\vspace{1.0ex},\\
    \mathbf{H}^{S_1} = \mathbf{W}^{M_1}\text{ReLU}(\mathbf{W}^{M_2} [\mathbf{H}^{S_1}_{P}, \mathbf{H}^{sg}_{C}]),
\end{array}
\label{eq3}
\end{equation}
where $\mathbf{H}_C$ indicates the fusion information of the compound proposition text and images, gained by the cross-modal encoder (arr. cross encoder as shown in Figure~\ref{fig:model}).
$\mathbf{W}^{M_1}\in\mathbb{R}^{2d\times d}$ and $\mathbf{W}^{M_2}\in\mathbb{R}^{2d\times 2d}$ are learnable parameters. Before feeding $\mathbf{H}(p)$ into the transformer, we introduce the learnable position embeddings $PE$ to facilitate it pay attention to the contextual information across images. 
After obtaining the final reasoning state $\mathbf{H}^{S_1}= \mathbf{(h_{1}^{+}, ..., h_{M}^{+})}$ of simple propositions in System 1, we adopt a linear prediction head to produce the confidence score of each proposition to images, which are defined as $P^{S_1}={(p^{+}_{1}, ..., p^{+}_{M})}$ and $p^{+}_{M}\in\mathbb{R}^{1\times L}$.

\subsection{System 2: Neural-Symbolic Reasoner}
\begin{figure}[t]
    \small
    \centering
    \includegraphics[width=0.38\textwidth, height=0.33\textwidth]{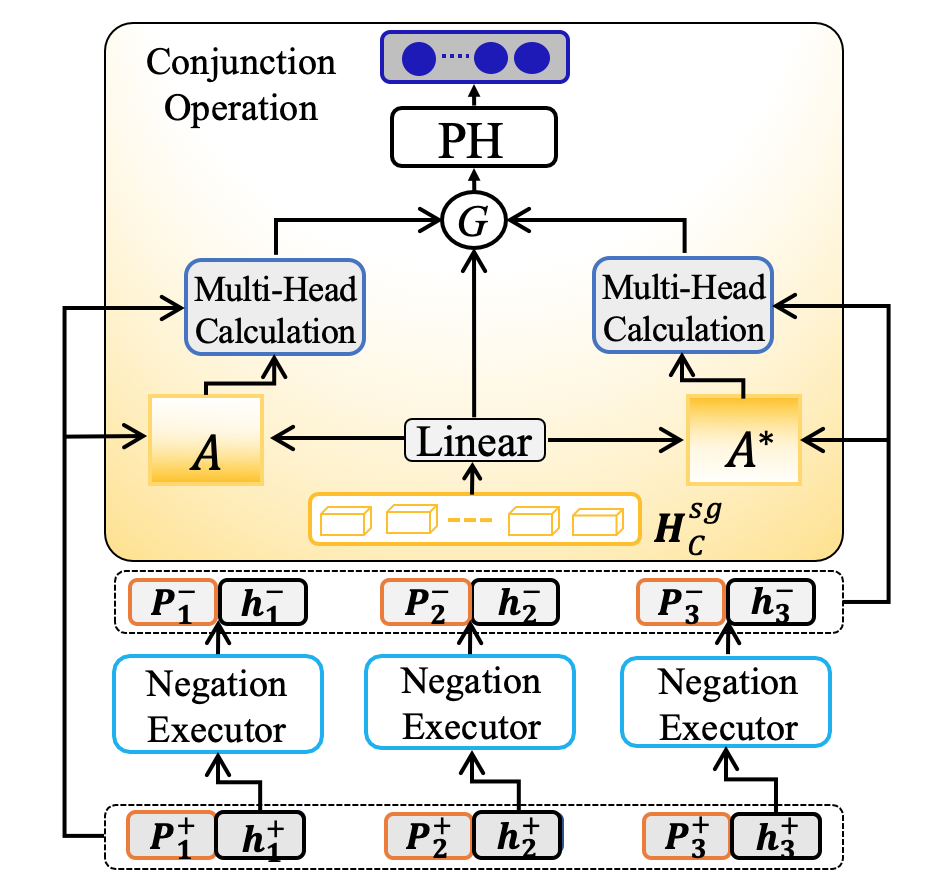}
    \caption{The detailed workflow of Neural-Symbolic Reasoner. It contains the underlying negation executor and upper conjunction operation.}
    \label{fig:symbol}
\end{figure}

For complex reasoning problems, the logical reasoning process usually plays a more significant role for intelligent machines and human reasoning~\citep{system_1_2}, which the visual-linguistic interactor is not capable of. Instead of combining the inferring results in System 1 via rule-based methods such as mean pooling, inspired by \citet{symbolic_neural, chen2021neural}, we devise a learnable Neural-Symbolic Reasoner~(NSR) to perform logical reasoning based on System 1 as shown in Figure~\ref{fig:model}. As depicted in Figure~\ref{fig:symbol}, it contains a negation executor to obtain the negational reasoning states and a conjunction operation to acquire the result of logical reasoning with attention to the positive and negational reasoning information.

\textbf{Negation Executor.} The negation executor is a module that takes the reasoning state of a simple proposition as input and produces the corresponding reasoning state of its negation as output. Its aim is to obtain useful cross-modal reasoning states for the negation of a proposition.
We regard $\mathbf{H}^{S_1}$ as the positive reasoning state and use a two-layer MLP with the ReLU activation function to obtain the negational reasoning state. The calculation process is given in Eq.~\ref{eq5},
\begin{equation}
    \text{NEG}(\mathbf{H}^{S_1}) = W^{n}_{2}\text{ReLU}(W^{n}_{1} \mathbf{H}^{S_1} + b_1^{n})  + b_2^{n},
\label{eq5}
\end{equation}
where $ W^{n}_{2}, W^{n}_{1}\in\mathbb{R}^{d\times d}$, $b_1^{n}, b_2^{n} \in\mathbb{R}^{1 \times d}$ are learnable parameters. We define the output of negation executor to $\mathbf{H}^{N} = (\mathbf{h^{-}_{1}}, ...,\mathbf{h^{-}_{M}})$, contrast to $\mathbf{H}^{S_1}$.The negational proposition has a different cross-modal reasoning state $\mathbf{H}^{N}$ than the corresponding positive proposition $\mathbf{H}^{S_1}$. We use the same linear prediction head as System 1 to produce the corresponding confidence score on images, which are presented to $P^{N} = (p^{-}_{1}, ..., p^{-}_{M})$. To make the negation executor effective, we will define a negational feedback loss to locally optimize it.


\textbf{Conjunction Operation.} 
Firstly, we define a new joint representation that incorporates reasoning hidden states and corresponding confidence scores as the initial state of conjunction operation. The process is presented in Eq.~\ref{eq4},
\begin{equation}
\begin{array}{c}
    \mathbf{P}^{+}_{i} = \text{Softmax}(p^{+}_{i})\cdot \mathbf{H}_{I},  i=1,...,M \vspace{1.1ex}, \\
    \mathbf{H}^{\textit{ns}}_{p_i^{+}} = [ \mathbf{P}^{+}_{i}, \mathbf{h_i^{+}} ], i=1,...,M,
\end{array}
\label{eq4}
\end{equation}
where $\text{[} , \text{]}$ indicates the concat calculation and $\mathbf{H}_{I}$ is the representation of images. $\mathbf{H}^{\textit{ns}}_{p_i^{+}}$ represents the positive joint representation of $i$ th proposition. We use the same calculation method as Eq.~\ref{eq4} to obtain the initialized negational representation $\mathbf{H}^{\textit{ns}}_{p_i^{-}}$. Then, we utilize the reasoning state of compound proposition text $\mathbf{H}^{sg}_C$ (Eq.~\ref{eq3}) as the signal to drive the conjunction calculation via the method of multi-head attention equipped with gate fusion, as shown in Figure~\ref{fig:symbol}. The whole calculation process is presented in Eq.~\ref{eq6},
\begin{equation}
\begin{array}{c}
    \mathbf{H}^{+} = \text{MultiHead}(W^{s}\mathbf{H}^{sg}_C, \mathbf{H}^{\textit{ns}}_{p_i^{+}}),\vspace{1.0ex} \\
    \mathbf{H}^{-} = \text{MultiHead}(W^{s}\mathbf{H}^{sg}_C, \mathbf{H}^{\textit{ns}}_{p_i^{-}}), \vspace{1.0ex}\\
    g^{+} = W^g[\mathbf{H}^{+}, W^{s}\mathbf{H}^{sg}_C] + b^g,\vspace{1.0ex}\\
    g^{-} = W^g[\mathbf{H}^{-}, W^{s}\mathbf{H}^{sg}_C] + b^g,\vspace{1.0ex}\\
    \mathbf{H}^{f} = W^{S_2}(g^{+}\mathbf{H}^{+} +  g^{-}\mathbf{H}^{-}),\\
\end{array}
\label{eq6}
\end{equation}
where $W^{s}\in\mathbb{R}^{2d\times 2d}$, $W^{g}\in\mathbb{R}^{1 \times 4d}$, $W^{S_2}\in\mathbb{R}^{2d\times d}$ are the learnable parameters and $\mathbf{H}^{f}\in\mathbb{R}^{1\times L \times d}$. We also utilize another linear prediction head to obtain the final confidence score of neural-symbolic reasoner, which is defined as $P^{S_2}\in\mathbb{R}^{1\times L}$.  

\subsection{Combining System 1 and System 2}

In addition, we combine inferring confidence scores in System 1 and System 2 to obtain the final solution, achieving the complementarity of System 1 and System 2. First, we need to acquire the whole representation of $\mathbf{H}^{S_1}$ and $\mathbf{H}^{f}$ as follows:
\begin{equation}
\begin{array}{c}
    \mathbf{H}^{f}_{W} = (W^{l}\mathbf{H}^{f} + b^{l})^{T}\mathbf{H}^{f}, \vspace{1.0ex}\\
    \mathbf{H}^{S_1}_{W} = (W^{l}\mathbf{H}^{S_1} + b^{l})^{T}\mathbf{H}^{S_1},
\end{array}
\label{eq7}
\end{equation}
where $W^{l} \in \mathbb{R}^{d\times1}$, $b^{l}$ are learnable parameters. $\mathbf{H}^{S_1}_{W}=(\mathbf{h}^{+}_{w1}, ...,\mathbf{h}^{+}_{wM})\in\mathbb{R}^{M\times d}$ and $\mathbf{H}^{f}_{W}\in \mathbb{R}^{1\times d}$ are used to gain the final solution via Eq.~\ref{eq8},
\begin{equation}
\begin{array}{c}
    \mathbf{h}_{c} = \sum\limits_{j=0}\limits^{M}{(W^{a}\mathbf{H}^{f}_{W} + W^{b}\mathbf{h}^{+}_{wj} + b^{c})}, \vspace{1.0ex}\\
    \hat{S}_{j} = V(W^{a}\mathbf{H}^{f}_{W} + W^{b}\mathbf{h}^{+}_{wj} + b^{c}) + b^{v},\vspace{1.0ex}\\
    sig = f(W^{f}[\mathbf{H}^{f}_{W}, \mathbf{h}_{c}] + b^{f}),\vspace{1.0ex}\\
    P^{f} = sig\cdot(\sum\limits_{j=0}\limits^{M}{\hat{S}_{j}p^{+}_{j}}) + (1-sig)\cdot P^{S_2},
\end{array}
\label{eq8}
\end{equation}
where $W^{a},W^{b}\in\mathbb{R}^{d\times d}$, $b^{c}\in\mathbb{R}^{d}$, $V\in\mathbb{R}^{d\times 1}$, $W^{f}\in\mathbb{R}^{2d\times d}$, $b^{v}, b^{f}\in\mathbb{R}^{1}$ are learnable parameters and $f(.)$ indicates the sigmoid activation function. This way, we can obtain the final result via taking the maximum one of the confidence score $P^{f}\in\mathbb{R}^{1\times L}$.

\subsection{Training Strategies}
\label{training_m}
To make the proposition generator perform proposition decomposition and generation effectively, we train it on a large-scale corpus solely and then train the whole NDCR framework on the specific training data. The two training phases are as follows:

\noindent\textbf{Phase 1.} We first pretrain the proposition generator on the released large-scale complex text simplification data set MinWikiSplit~\citep{minwikisplit}, which is composed of 203K pairs of aligned complex source and simplified target sentences. We adopt the cross entropy generation loss $\mathcal{L}_{g}$ for the decoder output. Similar to {S}im{CSE}~\citep{gao-etal-2021-simcse}, we employ the contrastive learning loss $\mathcal{L}_{c}$ to make the global representation of simple proposition sentence different. In addition, we use a cross-entropy multi-label classification loss $\mathcal{L}_{p}$ to train the prediction head of numbers of propositions, where the label is the number of simple sentences in the pretraining corpus. The whole training loss:
\begin{equation}
     \mathcal{L}_{phrase 1} =  \mathcal{L}_{g} + \mathcal{L}_{c} +  \mathcal{L}_{p}.
\end{equation}

\noindent\textbf{Phase 2.} While training NDCR, we employ the proposition sentence-image confidence score to calculate the classification loss. The loss will cover the output of System 1, System 2 and final solution, which is defined as follows:
\begin{equation}
     \mathcal{L}_{match} = \sum\limits_{i=0}\limits^{M+2}\text{cross-entropy}(p_i, q),
\end{equation}
where $p_i\in\mathbb{R}^{1\times L}$ and $q$ is the golden label. To make the negation executor effective, we devise a negational feedback loss $\mathcal{L}_{neg}$ to optimize it. We take the prediction result of modifier in System 1 as the positive distribution and make the belief distribution output by the negation executor on the image candidates be far away from positive distribution. The loss calculation method is shown in Eq.~\ref{eq11},
\begin{equation}
     \mathcal{L}_{neg} = \sum\limits_{z=0}\limits^{M}\text{max}(\theta - \text{KL}(p^{-}_z, p^{+}_z), 0.0),
\label{eq11}
\end{equation}
where KL indicates the K-L Divergence~\citep{kullback1951information}. $\theta$ is a super-parameter used to expand the positive and negational interval, which is set to 0.2. Hence, the whole optimization target is $ \mathcal{L}_{match} + \mathcal{L}_{neg}$.

\section{Experiments}
\subsection{Dataset}
We conduct extensive experiments on a challenging data set I{\footnotesize MAGE}C{\footnotesize O}D{\footnotesize E}~\citep{imagescode}, which contains 94,020 images, and they are divided into 9,402 sets. The overall images are collected from four released data sets: MSR-VTT~\citep{msrvtt}, Video-Storytelling~\citep{video_story}, YouCook~\citep{youcook}, and Open Images V6~\citep{openimages}. It consists of 21,202 human-writing complex descriptions and manually labelling corresponding golden images, which are divided into 16,594, 2,302, and 2,306 for training, validating, and testing, respectively. The image sources in the overall data set include video frames and static images.

\subsection{Baselines}
We compare NDCR with various types of pretrained VLMs and other designed models based on the specific condition of this task. Specifically, ViLBERT~\citep{vilbert} is a cross encoder where language and vision interact in the transitional layer via cross attention calculation. CLIP~\citep{CLIP} is a two-stream vision-language encoder with two independent visual and textual encoders. UNITER~\citep{chen2020uniter} is a single-stream encoder where visual representations and text tokens are concatenated and interact via the same transformer. OFA~\citep{ofa} is a unified cross-modal and unimodal encoder and has achieved impressive performance on multiple cross modal reasoning tasks. \citet{imagescode} also designed a contextual module to improve the interaction across different text-image fusion representations, achieving state-of-the-art performance.

\subsection{Implementation Details}
The $L$, $\lambda$, and $d$ equal 10, 1000, and 512, respectively. For the proposition generator, we adopt a two-layer semantic parsing module and the pretrained parameters of BART-base version. We set the maximum number of propositions to 10 and trained the proposition generator for 15 epochs on the MinWikiSplit data set. In addition, we set the depth of transformer block to 2 in the visual-linguistic interactor and utilized the finetuned visual encoder of CLIP (ViT-B/16) to encode images. For the cross encoder, we adopt the OFA-large architecture and first finetune it for two epochs before training the overall structure of NDCR. We froze the cross encoder, proposition generator, and visual encoder to prevent overfitting while training NDCR. While training all models, we set the batch size, initial learning rate, and dropout rate to $36$, $6\times1e^{-5}$, and $0.1$, respectively. The maximum training epoch is set to 30, and we employ the Adam Optimizer~\citep{kingma2014adam} with the initial learning rate declining linearly to train all models. We use the validation set to select the best-performing model.

\subsection{Main Results}
\begin{table}[t]
\renewcommand\arraystretch{1.10}
\begin{center}
    \scalebox{0.82}{
    \begin{tabular}{c c c c }
        \toprule
        \textbf{Method} $\downarrow$ \textbf{Type} $\rightarrow$ & All & Video & Static  \\
        \midrule
        CLIP~\citep{CLIP} & 28.4 & 20.0 & \underline{60.0} \\
        CLIP$^{\dag}$~\citep{imagescode} & \underline{29.9} & \underline{22.0} & 59.8 \\
        \midrule
        UNITER~\citep{chen2020uniter} & 24.8 & 17.4 & 52.8 \\
        UNITER$^{\dag}$~\citep{imagescode} & 25.7 & 19.1 & 50.5 \\
        \midrule
        ViLBERT~\citep{vilbert} & 20.9 & 15.0 & 42.7 \\
        ViLBERT$^{\dag}$~\citep{imagescode} & 24.5 & 18.0 & 49.3 \\
        \midrule
        NDCR~(ours) & \textbf{34.1} & \textbf{26.1} & \textbf{64.3}\\
        \bottomrule
    \end{tabular}}
    \caption{\label{tab:main_result} Model performance (accuracy) on \textbf{original testing set}. The results of CLIP, UNITER, ViLBERT, and their variants($\dag$) are reported by \citet{imagescode}. The underscore and bold indicate the second highest value and best performance (same as following tables). We report results for all examples and two disjoint subsets: video frames and static images.}
\end{center}
\end{table}

\textbf{Overall Performance.} We present the performance of NDCR and comparative models on the test set in Table~\ref{tab:main_result}. '$\dag$' indicates that the pretrained VLMs are equipped with the contextual module and temporal embedding to enhance the contextual semantic interaction across similar images. This variant shows its effectiveness on the case of video frame according to the comparative performances such as CLIP vs. CLIP$^{\dag}$.  Table~\ref{tab:main_result} reports that the proposed method achieves new state-of-the-art performance on the whole test set and significantly outperforms previous strong baseline ($34.1$ vs. $29.9$, $\uparrow$ 4.2). NDCR improves performances both on video frames and static images, especially static images($\uparrow$ 4.3), which shows its generalization on different cases. We observe that all models perform poorly on the testing samples whose images are from the video clips, which may be attributed to the high similarity across video frames. Hence, there is a big room to improve the whole performance on the challenging multi-modal reasoning task.

\subsection{Ablation Study}

\begin{table}[t]
\renewcommand\arraystretch{1.10}
\begin{center}
    \scalebox{0.82}{
    \begin{tabular}{c c c c }
        \toprule
        \textbf{Method} $\downarrow$ \textbf{Type} $\rightarrow$ & All & Video & Static  \\
        \midrule
        OFA~\citep{ofa}& 29.0	& 22.1	& 54.8\\
        OFA$^{\dag}$ &\underline{30.0}	& \underline{23.6}	& 54.6\\
        CLIP~\citep{CLIP} & 27.4	& 19.7	& \underline{56.5} \\
        CLIP$^{\dag}$~\citep{imagescode} & 27.6	& 20.8 & 53.2 \\
        \midrule
        NDCR~(ours) & \textbf{32.8} & \textbf{25.7} & 59.2\\
        \midrule
        System 2 & 32.4	& 25.3	& \textbf{59.3}\\
        System 2 w/o Negation & 32.0& 25.3	& 57.3 \\
        System 1 & 31.6	& 24.5& 58.3\\
        System 1 w/o Modifier & 19.3& 16.4	& 30.3\\
        \bottomrule
    \end{tabular}}
    \caption{\label{tab:ablation_study} Ablation experiments on the \textbf{testing$^{*}$} set, where we manually label the testing set to conduct ablation studies. 'Negation' and 'Modifier' indicate the negation executor and modifier. We adopt the mean pooling method to aggregate the predicted results of simple proposition in System 1 and w/o Modifier.}
\end{center}
\end{table}

\textbf{Effectiveness of Modules.} 
To study the effectiveness of different modules, we re-annotate the test sample with the help of eight related workers (original test labels are not released). The experimental results are presented in Table~\ref{tab:ablation_study}. The performances of reproduced baselines and NDCR have a slight decline, which is because the labelling process for most examples is difficult. There are specific quality differences across human-labelling results, yet it does not affect testing and comparing model performances. For the fairness of model comparison, the random seeds of all ablation experiments are set to the same value 10. Firstly, NDCR achieves the best performance and significantly surpasses other models on two test sets. When we add System 2 based on System 1, the overall performance improves by about 1.0, suggesting the neural-symbolic reasoner's effectiveness. Comparing System 2 and System 2 w/o negation, we observe that the negation executor improves the performance of the neural-symbolic reasoner, mainly in the case of static images. In addition, comparing System 1 and System 1 w/o modifier, we observe that introducing the context reasoning information is a very useful way to enhance the reasoning state representation of decomposed simple proposition sentences. Compared to the best baseline OFA-large (470M), the total parameter size of NDCR is about 440M.  NDCR has fewer parameters yet significantly outperforms it (as shown in Table 2). This suggests that the overall performance improvement of NDCR is not due to having larger parameters.

\begin{table}[t]
\renewcommand\arraystretch{1.10}
\begin{center}
    \scalebox{0.75}{
    \begin{tabular}{c c c c c c }
        \toprule
        \textbf{Method} $\downarrow$ \textbf{Nums\_of\_props} $\rightarrow$ & 1 & 2 & 3 & 4 & 5  \\
        \midrule
        Total Number & 61 & 863 & 1239 & 126 & 16\\
        \midrule
        CLIP$^{\dag}$ & 27 & 264 & 302 & 41 & 3\\
        OFA$^{\dag}$ & 28 & 299 & 340 & 34 & \textbf{5}\\
        System 1 & 30 & 297 & 364 & 36 & 4\\
        System 2 & \textbf{31} & \textbf{304} & 373 & 36 & 3\\
        NDCR & \textbf{31} & \textbf{304} & \textbf{380} & \textbf{37} & 3\\
        \midrule
        $\Delta$  & 3 & 5 & 40 & 2 & -2\\
        \bottomrule
    \end{tabular}}
    \caption{\label{tab:statistic} The number of samples accurately inferred on different numbers of simple proposition sentences. 'Nums\_of\_props' indicates the number of simple propositions. $\Delta$ represents the difference in the number of samples that NDCR and OFA$^{\dag}$ accurately predict.}
\end{center}
\end{table}

\noindent\textbf{System 1 vs. System 2.}
We count the experimental results on the test set according to the number of simple proposition sentences into which compound proposition texts are divided. The results are shown in Table~\ref{tab:statistic}. The statistical results show that NDCR excels at image retrieval from complex text with medium length, especially for those containing three simple proposition sentences. It verifies the proposed method's effectiveness in handling the complex image-text reasoning problem. Compared to System 1, System 2 performs better on test samples containing 2 or 3 simple proposition sentences, which suggests that the neural-symbolic reasoner can improve the conjunction operation of prediction results of decomposed propositions compared to rule-based methods such as mean pooling.

\begin{figure}[t]
    \centering
    \includegraphics[width=0.48\textwidth]{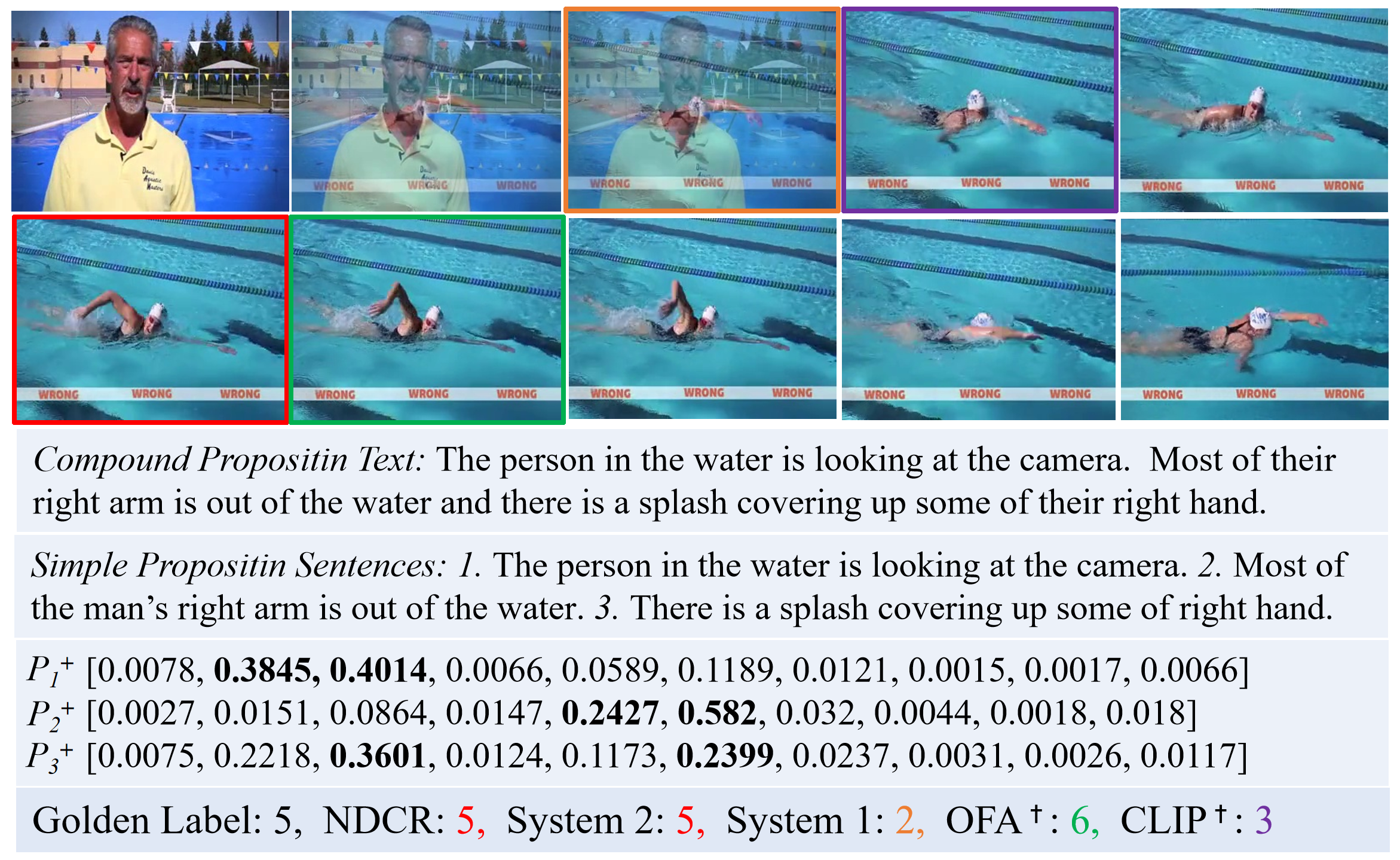}
    \caption{A case from the test set, where different colors correspond to the predicted result of models. $P^{+}_{1,2,3}$ represent the inferred confidence scores of simple proposition sentences in System 1 and are used to obtain the results in System 2 and final combination process.}
    \label{fig:case_1}
\end{figure}
\begin{figure}[t]
    \centering
    \includegraphics[width=0.48\textwidth]{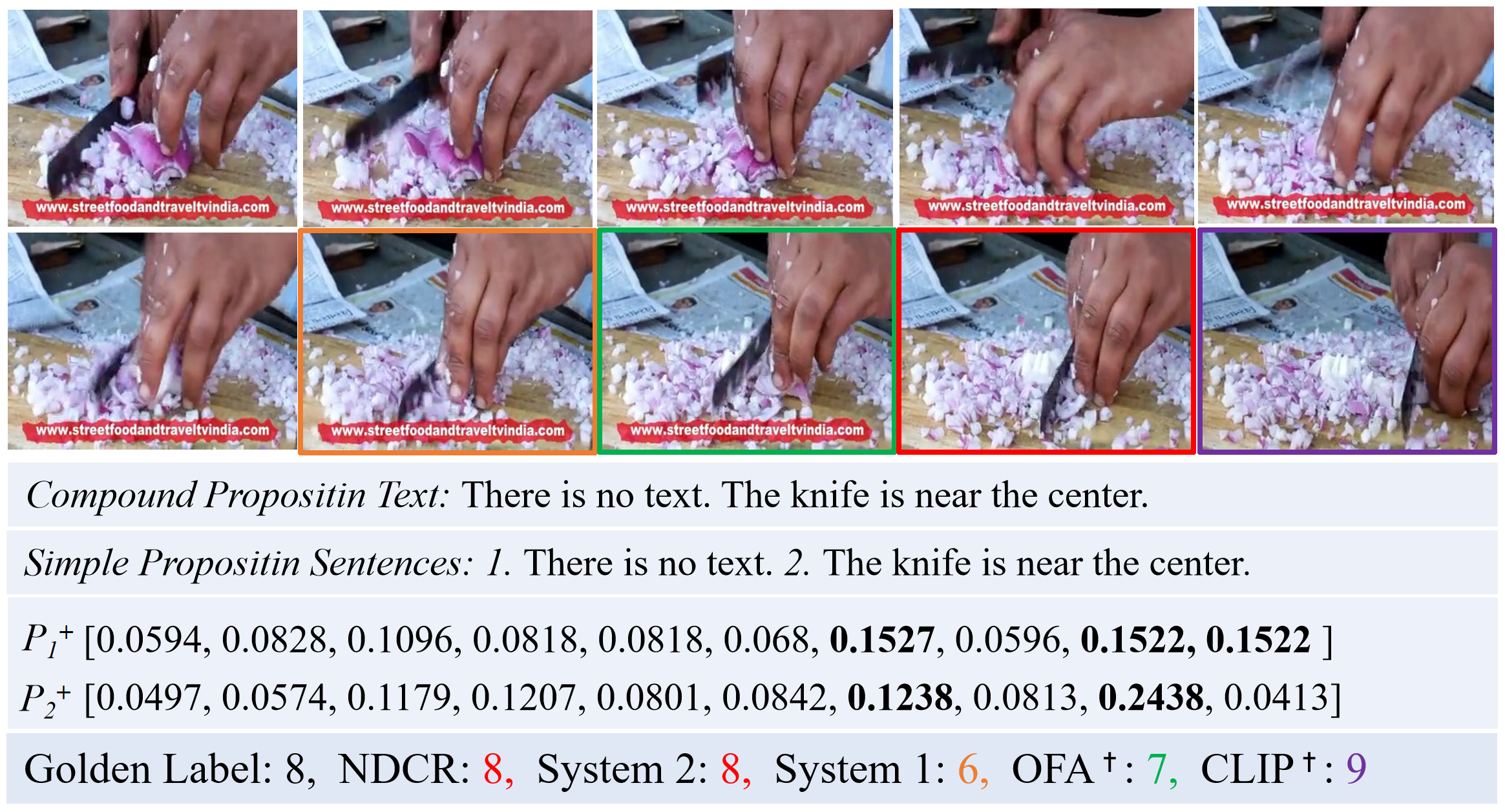}
    \caption{Another case from the test set, where it contains two simple proposition sentences.}
    \label{fig:case_2}
\end{figure}

\subsection{Case Study}
We present two cases in Figure~\ref{fig:case_1} and~\ref{fig:case_2}. For the first case (Figure~\ref{fig:case_1}), the proposition generator divides the complex text into three proposition sentences, and System 1 inferred the confidence scores ($P^{+}_{1,2,3}$) of them to ten images. Although these results of simple proposition sentences contain some errors due to having no explicit supervision signal to train, System 2 (neural-symbolic reasoner) could obtain the correct result with logical reasoning operation compared to the rule-based aggregation method in System 1. It indicates the robustness of System 2. In addition, we observe that the pretrained VLMs and System 1, which are capable of perceptual computing, often fail to cover all text semantics. It is easy for them to ignore pivotal text information (such as ``there is no text'' shown in Figure~\ref{fig:case_2}), which leads to inference errors. In conclusion, combining logical reasoning System 2 and powerful analogical reasoning System 1 (e.g., pretrained VLMs) has significant potential to take their advantages to address complex reasoning problems.

\section{Conclusion}

In this paper, inspired by the divide-and-conquer algorithm and dual-process theory, we introduced an end-to-end neural divide-and-conquer reasoning framework named NDCR to handle the challenging case of image retrievals from linguistically complex text. NDCR contains a proposition generator to divide the compound proposition text into multiple simple proposition sentences, then uses a visual-linguistic interactor to achieve the interaction of simple propositions and images. To improve the logical reasoning capability, we devise a neural-symbolic reasoner to gain the logical inferring result based on the output of the visual-linguistic interactor. This way, NDCR performs the low-level analogically perceptual computing in System 1 (visual-linguistic interactor) and high-level logical reasoning in System 2 (neural-symbolic reasoner). Finally, we combine the output result in Systems 1 and 2 to obtain the final solution. 

\section*{Limitations}

The proposed method NDCR has some limitations as follows:
1) The produced representation of simple proposition sentences in the proposition generator lies in a different space distribution with the image encoding, which affects the performance of their fused representation.  Although we introduce the reasoning information of compound proposition text to alleviate this issue, we hope to solve it by improving the text understanding capability of pretrained VLMs. In addition, adopting the pretrained textual encoder of VLMs to perform proposition decomposition is inadequate due to that they present an inferior understanding for the discourse structure of long texts.
2) The performance of samples with highly similar images from video frames is quite different from that of humans. We may improve it from the perspective of image difference modelling.
3) The experimental results indicate that our method is effective at logical inference on examples with medium-length descriptions, but there is still room for improvement for longer descriptions.

\section*{Ethics Statement}
I{\footnotesize MAGE}C{\footnotesize O}D{\footnotesize E}~\citep{imagescode} is an open data set used for scientific research. For ablation studies in the test set, we hired masters and undergraduate students from the research group to re-annotate the label of the test set. We have informed the creators of the data set and only conducted scientific research.

\bibliography{custom}
\bibliographystyle{acl_natbib}


\end{document}